\documentclass[10pt,journal,comsoc]{IEEEtran}

\usepackage{graphicx}
\usepackage{longtable}
\usepackage{textcomp}
\usepackage{amsmath,amssymb,amsfonts}
\usepackage{xcolor}
\usepackage{graphics}
\graphicspath{ {c:/user/images/} }
\usepackage{tikz}% Required for inserting images
\usepackage{blindtext}
\usepackage{longtable}
\usepackage{authblk}
\author[1]{Sitara Afzal}
\author[1]{Haseeb Ali Khan}
\author[1]{Imran Ullah Khan}
\author[2,*]{Md. Jalil Piran}
\author[1,*]{Jong Weon Lee}
\affil[1]{Software Department, Sejong University, Seoul 05006, South Korea}
\affil[2]{Department of Computer Science and Engineering, Sejong University, Seoul 05006, South Korea}
 \affil[*]{Corresponding author:  jwlee@sejong.ac.kr, piran@sejong.ac.kr}

\begin{document}
\title{A Comprehensive Survey on Affective Computing; Challenges, Trends, Applications, and Future Directions}
\maketitle
%\date{March 2023}
\begin{abstract}
As the name suggests, affective computing aims to recognize human emotions, sentiments, and feelings. There is a wide range of fields that study affective computing, including languages, sociology, psychology, computer science, and physiology. However, no research has ever been done to determine how machine learning (ML) and mixed reality (XR) interact together. This paper discusses the significance of affective computing, as well as its ideas, conceptions, methods, and outcomes. By using approaches of ML and XR, we survey and discuss recent methodologies in affective computing. We survey the state-of-the-art approaches along with current affective data resources. Further, we discuss various applications where affective computing has a significant impact, which will aid future scholars in gaining a better understanding of its significance and practical relevance.
\end{abstract}

\section{Introduction}
Emotions, sentiments, and feelings, along with emotion recognition, constitute what is referred to as "affective computing" \cite{ref1}. In 1997, Prof. Picard introduced the idea of affective computing, which has helped computers recognize and communicate their moods, as well as effectively respond to humans' moods \cite{2} \cite{3}. In many practical applications, it is desirable to develop a cognitive, intelligent system that can detect and comprehend people's feelings while also providing sensitive and cordial responses \cite{4}. Emotion is a cultural and psychobiological adaptation mechanism that enables people to adapt dynamically to environmental changes. Emotions give meaning to our lives, deepen our relationships with others, alert us to our wants and sentiments, and help us make changes \cite{6}. 

There are five components to a single sentiment: a cognitive evaluation, a physical sensation, an intention, a subjective "feeling," a motor reaction, and, in most situations, an interpersonal component. "Emotion Regulation" (ER) is a group of mental processes that affect our emotions, as well as the timing of our emotions \cite{8}. A dynamic process intended to downregulate or upregulate positive or negative thoughts, it is essential to human mental function. In order to identify expressed emotions (anger, disgust, fear, happiness, sorrow, or surprise) and predict their sentiment (positive or negative), a framework based on frame of reference intermodal attention \cite{9} was developed. 

It is called emotion dysregulation when a person is unable to effectively control or process their emotions, resulting in unintended intensification or deactivation \cite{3}. Managing emotionally intense experiences is therefore a component of self-regulation. A specific method of emotion management is not necessarily beneficial or harmful \cite{10}. In doing so, it avoids making generalized judgments about which coping mechanisms are more or less adaptive depending on the situation \cite{11} \cite{12}.  In various social media platforms, affective computing can greatly help in understanding the thoughts expressed \cite{9}. In this way, emotional computing is seen by many researchers as a way to advance creation of human-centric AI and human intelligence \cite{13}. There is a difference between sentiment research and emotion recognition in affective computing \cite{14}. A diagram showing the different kinds of emotions can be found in figure 1.

\begin{figure*}[t]
    \centering
    \includegraphics[width=0.5\textwidth]{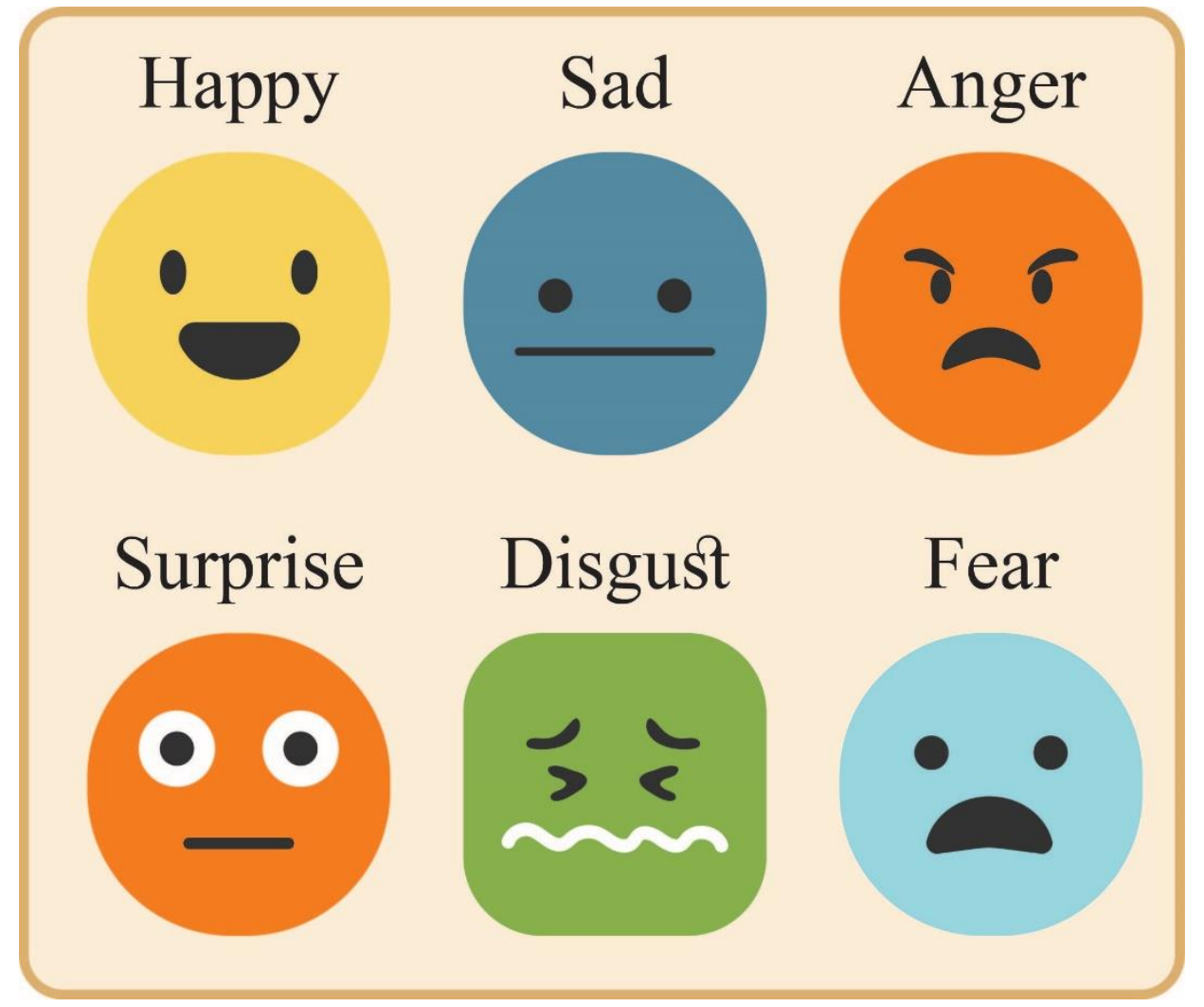}
    \caption{Types of Emotions}
    \label{fig1}
\end{figure*}

To understand and calculate human feelings, psychologists presented two common theories: the discrete emotion model (or categorical emotion model) \cite{15} and the dimensional emotion model \cite{16}. In order to identify people's emotional state, three main areas of focus for emotion recognition have been investigated \cite{17}: visual emotion recognition (VER), audio/speech emotion recognition (AER/SER) and physiological emotion recognition (PER). It has been shown that VR systems can elicit emotional responses that lead to psychologically positive changes when experiencing increased presence in a virtual environment \cite{21}. These widely used datasets have inspired the development of emotional computing, which is Machine Learning (ML). Study \cite{22} reports that in everyday human interaction, human emotions are primarily expressed through facial expressions (55\%) voice (38\%) and words (7\%) Textual, auditory, and visual signals are generally referred to as physical data. 

People's propensity to openly express their ideas and thoughts on social media platforms and websites makes it easy to gather a wide range of bodily affect information. As a result of these data, many academics attempt to identify minor emotions conveyed either explicitly or implicitly. As a result, \cite{120} and \cite{121} present an innovative approach to multimodal music emotion analysis in affective computing using LSTM networks. With the proposed model, a dual-channel LSTM replicates the human auditory and visual processing pathways, enabling the emotional information of music and facial expressions to be effectively processed. As a result, tangible affect recognition may not be successful, since people may intentionally or unintentionally cover up their feelings (called social masking) \cite{19}. 

\subsection{Motivation}
Increasingly, scientists in the fields of ergonomics and intelligent systems are working to improve the effectiveness and adaptability of human-computer interaction (HCI) in various situations. Computers' ability to comprehend human emotions and behavior is a crucial component of their flexibility. The performance of the state-of-the-art methods and their implications for recognition have not been fully addressed in previous surveys. The most important contribution of our review is that it addresses all areas of emotional computers through a variety of research methods and findings, as well as discussions and future efforts. Currently, most HCI technologies cannot recognize human emotion, but the development of advanced HCI systems depends on automatic computer recognition of human emotion.
 
\subsection{Contributions}
The performance of state-of-the-art methods and their implications for their recognition ability have not been fully addressed in previous reviews. 
 Our review offers a variety of research methods and findings, as well as discussions and future efforts, to address all aspects of emotional computers. The major contributions of this paper are as follows.
 \begin{itemize}
\item To the best of our knowledge, this is the only survey of its kind to use ML to categorize emotional processing into broad categories of emotion identification.      
\item By examining how well the various affective modality is utilized to study and identify affect, we provide a comprehensive classification of the state-of-the-art emotional computational tools.
\item To the best of our knowledge, this is the only survey that classifies emotional processing using mixed reality machines and even further taxonomies them based on augmented and virtual reality. 
\item In order to categorize benchmark databases for emotional computing, audio, speech, text, and visual modalities are used. Those resources' salient features and accessibility are outlined. 
\item Current approaches and tools are discussed along with their limitations.
\item Finally, we outline several open research problems in the field of augmented reality affective computing and ML.
 \end{itemize}
 \begin{table}[t]
    \caption{List of acronyms}

\centering
    \begin{tabular}{p{60pt} p{150pt}}
         \hline
         \hline
         \textbf{Acronym}& \textbf{Definition} \\
         \hline
         VER& Visual Emotion Recognition \\
         SER& Speech emotion recognition\\
         AER&Audio emotion recognition\\
         VR&Virtual Reality\\
         AR&Augmented Reality\\
         XR&Mixed Reality\\
         ML&Machine learning\\
         SVM&Support vector machine\\
         KNN&K-Nearest Neighbor\\
         RF&Random Forest\\
         ANN&Artificial Neural network\\
         DT&Decision Tree\\
         ER&Emotion regulation\\
         HCI&Human Computer Interface\\
         FER&Facial emotion Recognition \\
         \hline
         \hline
    \end{tabular}
    \label{tab1}
\end{table}

\begin{figure*}[t]
    \centering
    \includegraphics[width=0.8\textwidth]{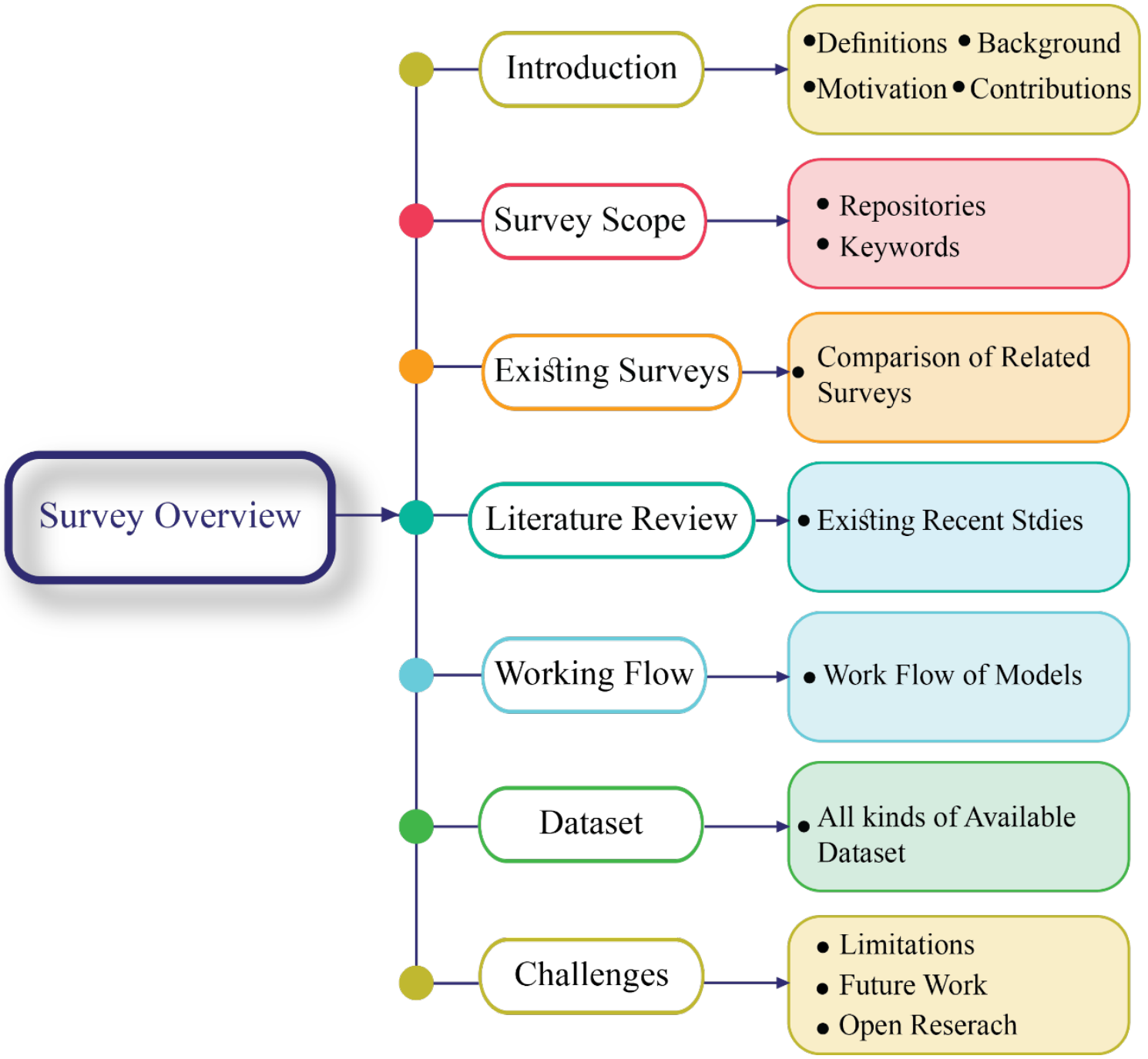}
    \caption{Organization of this survey paper.}
    \label{fig2}
\end{figure*}

The remainder of the paper is organized as follows. The scope of the paper is presented in Section II. The existing survey papers are discussed in Section III. In Section IV, a number of important techniques are compared. The dataset is covered in detail in Section V. In Section VI, the main challenges of the paper are summarized and several research directions are identified. Finally, Section V draws the conclusions. An overview framework is shown in Figure 2 to improve the general readability of the paper.

\section{Scope of this survey}

We will discuss the approach used in this literature review in this section. Affective computing as it pertains to ML, deep learning, and mixed reality is the focus of this survey, to the best of our knowledge. As far as we know, no other study has taken into account affective computing in terms of both ML and mixed reality. The relevant articles between 2014 and 2021 are considered. There are three main models in this survey: identification, screening, and inclusion.

As the first step of this survey, 'Identification' is used to classify the data from all the databases. To conduct this survey, we consider some research questions and keywords. VR, AR, affective computing, ML, supervised learning, and emotion recognition are explored and acquired from various publishers icnluding Institute of Electrical and Electronics Engineers (IEEE), Springer, Hindawi, and Association for Computing Machinery (ACM). 

We consider only the relevant articles from all the selected articles during this step. The duplicate record is removed using EndNote X9 for relevant articles that are not redundant. To determine whether the study articles' titles and abstracts were relevant to a developed research topic, the titles and abstracts were manually reviewed. We evaluated research papers that contributed a significant amount but were highly relevant.

Data selected following the qualifying stage are subjected to quantitative and qualitative evaluations at this stage of the research process. Adding records is the first step in the quality check analysis. During the last stage, known as qualitative analysis, some of these records are analyzed using meta-analysis. As a final step, data is extracted from such data that were included in the survey's final analysis.

\section{Related Work}

The key findings of this study are summarized in Table 2 in comparison with earlier research in the area of affective computing. In no previous study for emotion recognition, ML-based and mixed reality approaches have been examined concurrently. Prior research also ignored the numerous disciplines that contribute to the field of affective computing, as well as their ideas, models, and methodologies. This study discusses the classification and recognition of emotions in terms of their categories and methods for detecting them. We have also examined the challenges, remedies, and possible trends of research studies. 

Researchers provide a systematic review of emotion models, databases, and recent advances in \cite{24}, which includes both physical and psychological studies. In addition, they examined the topologies and capabilities of cutting-edge multimodal emotional analysis and unimodal affect identification. A mixed reality or ML approach to emotion recognition was not considered by the authors of \cite{25}. Affective computing is only considered in the context of interdisciplinary fields like psychology and computer science, along with their theories, concepts, models, and implications. Additionally, they present some existing affective databases.
In \cite{26}, authors surveyed the state-of the art studies <of ????>. Their main focus was on the types of features in emotions datasets as they reviewed affective image contents. In affective computing, they discuss approaches to identifying emotions using features extraction. The authors of \cite{27} described the state-of-the-art affective computing technologies for large-scale heterogeneous multimedia data in a comprehensive manner. Using handcrafted features-based features, they compare relevant techniques on AC of many multimedia forms, including photos, music, movies, and multimodal data.

\begin{table*}[t]
    \centering
    \caption{Summary of existing surveys.}
    \begin{tabular}{p{40pt}p{30pt} p{30pt} p{30pt} p{50pt} p{30pt} p{40pt} p{40pt} p{40pt}}
         \hline
         \hline
         \textbf{Research}&\textbf{Year}&\textbf{AR/VR}&\textbf{ML}&\textbf{All four categories
A/V/T/S}&\textbf{DB}&\textbf{Wearables}&\textbf{Accuracy}&\textbf{Applications} \\
         \hline
         \hline
        \cite{24} & 2022&
         $\times$ &
         \checkmark&
         \checkmark&
         \checkmark&
         $\times$&
         \checkmark&
         \checkmark\\
         \hline

         \cite{25} & 2021&
         $\times$ &
         $\times$&
         \checkmark&
         $\times$&
          $\times$&
           $\times$&
            \checkmark\\
         \hline
         
           \cite{26} & 2018&
         $\times$&
         \checkmark&
         $\times$&
         \checkmark&
         $\times$&
         $\times$&
         \checkmark\\
         \hline

        \cite{27} & 2019&
         $\times$ &
         \checkmark&
         \checkmark&
         \checkmark&
         $\times$&
         $\times$&
         $\times$\\
         \hline
         
       \cite{28} & 2018&
         $\times$ &
         \checkmark&
          $\times$&
          \checkmark&
         $\times$&
         \checkmark&
         $\times$\\
         \hline

      \cite{29} & 2018&
         $\times$ &
         \checkmark&
         $\times$&
         \checkmark&
         $\times$&
         \checkmark&
          \checkmark\\
         \hline

       \cite{30} & 2018&
         $\times$ &
         \checkmark&
         $\times$&
         \checkmark&
         $\times$&
         \checkmark&
          \checkmark\\
         \hline
          
        \cite{31} & 2020&
         $\times$ &
         \checkmark&
         $\times$&
         $\times$&
         $\times$&
         $\times$\\
         \hline

        \cite{32} & 2019&
         $\times$ &
         \checkmark&
         $\times$&
         \checkmark&
         $\times$&
         \checkmark&
         $\times$\\
         \hline

        \cite{33} & 2022&
       \checkmark&
         $\times$ &
         $\times$&
        $\times$&
         \checkmark&
         $\times$&
         \checkmark\\ 
         \hline

         Our Survey & 2023&
         \checkmark &
         \checkmark&
        \checkmark&
         \checkmark&
         \checkmark&
         \checkmark&
         \checkmark\\  
         \hline
         \hline
    \end{tabular}
    \label{tab6}
\end{table*}

Additionally, this study examines the different uses, problems, and upcoming challenges of emotion detection. In order to achieve the study's objective, the following research questions were developed:
\begin{itemize}
\item (RQ1) What different interconnected domains are a part of emotion detection?
\item (RQ2) What are the most common fields wherein affective computing is used?
\item (RQ3) Which recent research questions might have an impact on future work in emotion detection?
\item (RQ4) How ML and Mixed Reality recognizes the emotions.
\end{itemize}

A thorough search on the Microsoft academic research portal was conducted to obtain a comprehensive understanding of the published literature on affective computing.
\begin{figure*}[t]
    \centering
    \includegraphics[width=0.8\textwidth]{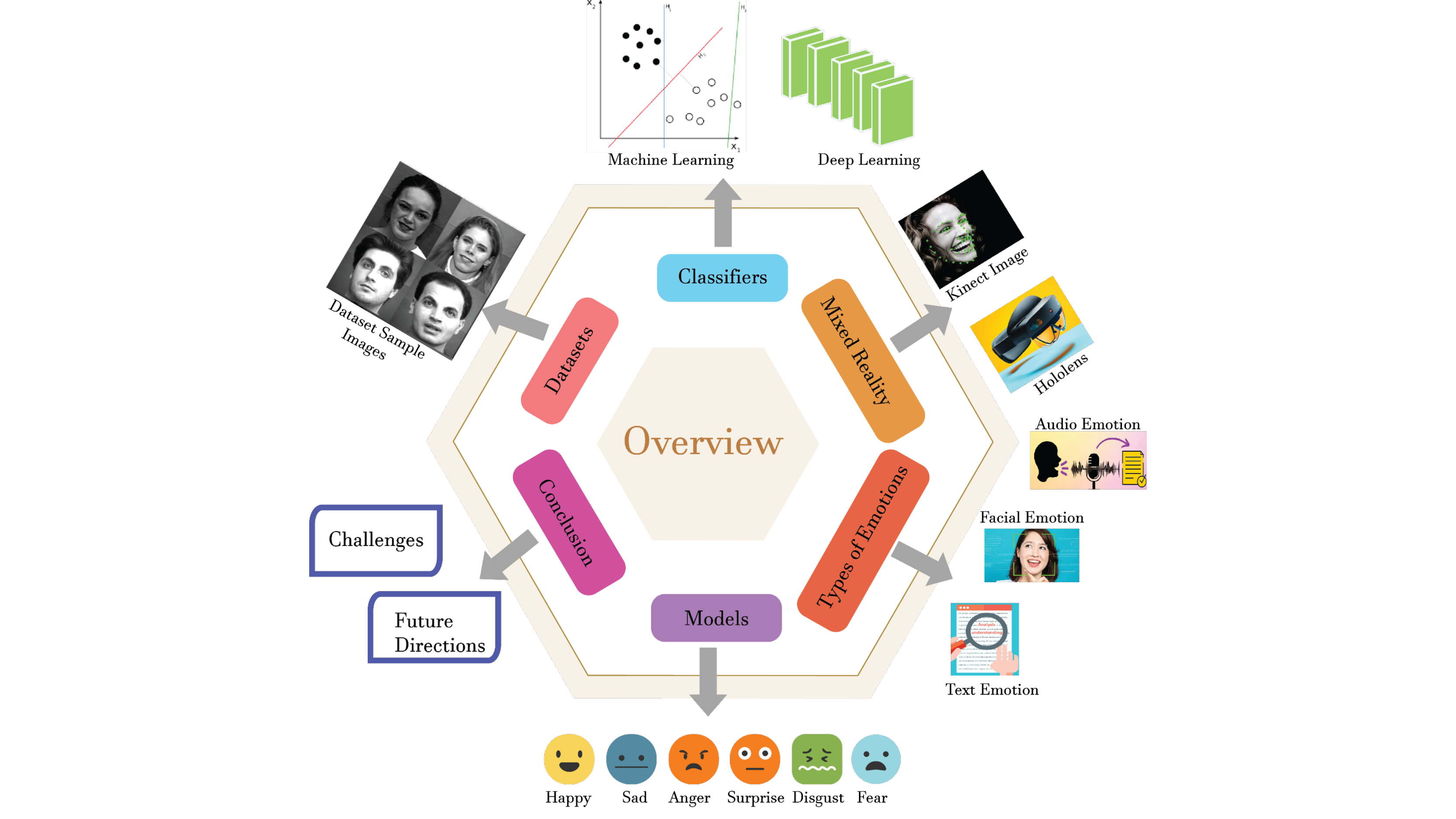}
    \caption{The overall framework of this survey}
    \label{fig3}
\end{figure*}

\section{Emotion Models}
The definition of the emotion or affect is crucial to building an emotional computing standard. Ekman articulated the fundamental idea of emotions for the first time in the 1970s. Despite psychologists' efforts to categorize emotions in neuroscience, philosophy, and computer science, there is no universally accepted emotion model. Generic emotion models used in affective computing include continuous emotion models and multidimensional emotion models (sometimes called continuous emotion models).

\section{Work Flow of Emotion Recognition}
In this section, we discuss the workflow and contributions of all forms of affective computing, including ML, deep learning, and virtual reality.

\subsection{Textual Based Emotion Recognition}
The majority of methods for textual emotion recognition, or TSA, are based on statistical or knowledge-based approaches [34]. In the former case, an extensive emotional vocabulary is required to model a thesaurus, while in the latter case, a large database with emotional labels is required. Online social media and e-commerce systems are rapidly expanding, allowing users to freely express their opinions, which generates a significant amount of text data. A text sentiment classification method was developed to detect subtle sentiments or emotions expressed directly or implicitly from user-generated material \cite{35}. 

The term "feature engineering" is often used to identify characteristics that are associated with emotion by using traditional approaches \cite{36}. Through the use of DL-based models, it is possible to create an end-to-end sentiment analysis of text data. Statistics and knowledge-based methods are mainly used in conventional ML-based TSAs [34]. It is common for semantically polarized terms to be mixed up in various lexicons, including Word-Net, WordNet Affect, Sentic-Net, and Senti-WordNet \cite{37}. Table 3 presents the articles and their contribution to text-based emotion detection using deep learning and ML techniques. Table 3 provides an overview of the main contributions along with datasets and evaluation parameters.

%TEXT BASE EMOTION TABLE
\begin{table*}[h]
    \centering
    \caption{An overview of text based emotion recognitions studies }
    \begin{tabular}{p{30pt} p{250pt} p{50pt} p{50pt}} 
        \hline
        \hline
  \textbf{Research}&\textbf{Contributions}&\textbf{Dataset}&\textbf{Evaluation} \\
        \hline
         
        \cite{38}& 
      Lexicon-based strategy, external evidences system, Context dependent approach

        &Review Dataset& Accuracy\\ 
        \hline
        
        \cite{39}& Baseline lexicons, Train word-class correlations, &Weblogs& Accuracy\\
        \hline
        
        \cite{40}& Concept-level sentiment analysis, logic computing, ML
&VER& Visual Emotion Recognition\\
        \hline
        
        \cite{41}& Multidisciplinary sentiment classification, Twitter ad hoc conversations of significant conflicts
&Twitter& Accuracy\\ 
        \hline
        
        \cite{42}& Asses sentimenatl analysis, context for lexicons methodologies
&Twitter& Accuracy\\ 
        \hline
        
        \cite{43}& Sentiment classification using SVMs, combination of phrases and adjective for favorability measures 
.&Music review& Accuracy\\ 
        \hline
        
        \cite{44}& The opinion-level framework, intra-opinion features and inter-opinion, Bayesian model &Twitter& Accuracy\\ 
        \hline
        
        \cite{45}& Strengthen the impartial, boundary between positive and negative reviews &Amazon, cinema, movies,& Accuracy\\ 
        \hline
        
        \cite{46}& Hybrid strategy, combination of deep learning and a vocabulary method &Amazon, IMBD and Yelp& Accuracy\\
        \hline
        
        \cite{47}& Identification of Internet slang and emoticons, k-Nearest Neighbors, Decision Tree, Random Forest, Logistic Regression, Naive Bayes, and Support Vector Machine
&Weibo& Accuracy\\
        \hline
        
        \cite{48}&	New word embedding technique, Dynamically adjust vector representation &IMDB Yelp 13, 14 and 15& Accuracy\\
        \hline
        
        \cite{49}& Novel text processing design (VDCNN), character level, tiny convolutions and pooling procedures &Amazon& Accuracy\\
        \hline
        
        \cite{50}& Unique parameterized convolutional neural network, Categorization of sentiments, Incorporate aspect data into CNN &SemEval 2014& Accuracy\\
        \hline
        
        \cite{51}&	LM employing contextual BLSTM (cBLSTM), Modified version of bidirectional LSTM (BLSTM)
&IMDB& Accuracy\\
        \hline
        
       \cite{52}& Joint framework for explicit aspect, Opinion terms co-extraction unifies recursive neural networks  &Laptop& Accuracy\\
        \hline
        
        \cite{53}& Examination of two distinct types of tweets obtained during the COVID19 pandemic &Twitter& Accuracy\\
        \hline
        \hline
        
        \end{tabular}\label{tab2}
\end{table*}

\subsection{Audio Emotion Recognition}
A process of processing and understanding speech signals is used in audio emotion recognition (also known as SER) \cite{54}. There are a number of ML- and DL-based SER systems that have been implemented to enhance analysis \cite{55, 56}. In ML-based SER, the extraction of acoustic features and the selection of classifiers are the primary focuses. A DL-based SER constructs a CNN architecture without taking feature engineering and selection into account, however, in order to forecast the final feeling \cite{57}.

In ML-based SER systems, two crucial processes take place: learning representations of emotional speech and selecting an appropriate classification for the final emotion prediction. A well-known audio feature extraction tool called OpenSMILE \cite{58} extracts all the essential components of speech. For SER systems, HMMs, GMMs, SVMs, RFs, and ANNs are frequently employed as classifiers. The SER also uses maximum classification and enhanced traditional easy-to-understand classifiers in addition to these learners. The following Table 4 shows an overview of the main contributions, along with the dataset and evaluation parameters.
%AUDIO BASE EMOTION TABLE
\begin{table*}[t]
    \caption{An overview of audio based emotion recognitions studies}
    \centering
    \begin{tabular}{p{40pt} p{250pt} p{50pt} p{50pt}}
        \hline
        \hline
        \textbf{Research}&\textbf{Main Contributions}&\textbf{Dataset}&\textbf{Evaluation} \\
        \hline
        
        \cite{59}& Examination of two types of tweets obtained during the COVID19 &Speaker language & Accuracy\\
        \hline        
         
        \cite{60}&Utilization of prosodic features, Utilization of voice quality parameters,  Investigation of the interactions between prosodic and voice quality, Utilization of the bayesian classifier to recognize emotions
&Speaker& Accuracy\\ 
        \hline
        
        \cite{61}& Fuse the features, multiple kernels learning-based approach 
&Berlin database& Accuracy\\ 
        \hline
        
        \cite{62}& Deep Neural Network based identification of emotions &Speaker& Accuracy\\
        \hline
        
        \cite{63}& Validate the use of a corpus of semantically impartial text, Transcription of both neutral and emotional (acted) speech
&TTS& Accuracy\\
        \hline
        
        \cite{64}&	Automated speech emotion detection algorithm, Computer model of the human auditory system &Speaker& Accuracy\\
        \hline
        
        \cite{65}& Classification on three corpora, namely the RAVDESS, IITKGP-SEHSC, and the Berlin EmoDB using SVM as classifier &Speaker& Accuracy\\
        \hline
        
        \cite{66}& Merging of personalized and non-personalized features for speech emotion, Utilze fuzzy C-means clustering algorithm, Employe several trees to identify various emotional states&CASIA& Accuracy\\
        \hline
        
        \cite{67}&Extraction of keypoints from spectrogram, Utilizion of CNN, Predictions of seven emotions &Speaker& Accuracy\\
        \hline
        
        \cite{68}& Performs auto encoders, Utilzioon of Recurrent neural networks, Classify four fundamental moods, Recognize emotions from spectrogram features
&Speaker& Accuracy\\
        
        \hline
        \hline
    \end{tabular}
    \label{tab3}
\end{table*}

\subsection{Visual Emotion Recognition}
Images or films with face emotional clues are used to execute FER \cite{69}. FER techniques are summarized in Table 5 as representative examples.  
%VISUAL BASE EMOTION TABLE
\begin{table*}[t]
    \caption{An overview of visual based emotion recognitions studies}

\centering
    \begin{tabular}{p{40pt} p{290pt} p{40pt} p{40pt} }
        \hline
        \hline
        \textbf{Research}&\textbf{Main Contributions}&\textbf{Dataset}&\textbf{Evaluation} \\
        \hline
         \cite{70}& Fully automatic face expression identification, Utilize subsequent frames based on elastic bunch graph, &CK+& Accuracy\\ 
        \hline
        
        \cite{71}& Fuzzification, Track AAM characteristics
&custom& Accuracy\\
        \hline
        
        \cite{72}& Linear discriminant analysis to recognize emotions &JAFFEE MMI SFEW& Visual Emotion Recognition\\
        \hline
        
        \cite{73}& Auto–encoders, SOM-based classification, Use of deep neural network, Combine geometric characteristics with LBP features
&MMI& Accuracy\\ 
        \hline
        
        \cite{74}& Auto–encoders, SOM-based classification, deep neural network, Combine geometric characteristics with LBP features
&CASME II& Accuracy\\ 
        \hline

        \cite{75}& Subtle deformations in emotion classification & BU-4DFE& Accuracy\\ 
        \hline
        
        \cite{76}& FER employing multi-model 2D and 3D films, convolution network, Encode both static and dynamic information
&BU-4DFE& Accuracy\\ 
        \hline
        
        \cite{77}& Recognition of face micro-expressions, weighting scheme as well as the excellent spatiotemporal descriptor HOG3D for action recognition
&HOG3D& Accuracy\\ 
        \hline
        
        \cite{78}& TCreation of the neutrality facial images &BU-4DFE CK+ MMI& Accuracy\\
        \hline
        
        \cite{79}& The "identification by generating" system, brand-new hard negative generation (HNG) network and a generalized radial metric learning (RML) network, is the main contribution of this research.
&CK+, MMI& Accuracy\\
        \hline
        
        \cite{80}& CNN to operate as distinct streams within a bi-stream identity-aware net &MMI CK+ & Accuracy\\
        \hline
        
        \cite{81}& Attention-based Salient Expressional Region Descriptor (SERD) and the Multi-Path Variation-Suppressing Network modules &FER& Accuracy\\
        \hline
        
        \cite{82}& Uncertainty using a straightforward but effective Self-Cure Network (SCN) &SemEval 2014& Accuracy\\
        \hline
        
        \cite{83}&	Dynamic-temporal stream, static-spatial stream, and local-spatial stream module for the TSCNN, Aims to learn and incorporate time, complete facial area, and nearest neighbor signals again for face
&CASME II, SMIC-HS, SAMM& Accuracy\\
        \hline
        
        \cite{84}& Deep emotion-conditional adaptation net (ECAN) that can learn domain-invariant and discriminative feature representations
&CK+ MMI JAFFE& Accuracy\\
        \hline
        
        \cite{85}& Transfer knowledge of Emotional Education Mechanism (EEM), Self-taught student network and a competent teacher network (STSN)
&Twitter& Accuracy\\
        \hline
        
        \cite{86}& Convolutional networks to encode the spatial properties of microexpressions at several expression-states &CASME-II& Accuracy\\
        \hline
        
        \cite{87}& Siamese-cascaded metric learning framework that teaches fine-grained distinctions between expressions in video-based tasks &CK+, Oulu, MMI& Accuracy\\
        \hline
        \hline
    \end{tabular}
    \label{tab4}
\end{table*}

\subsection{Virtual Reality Based Emotion Detection}
Emotional representation can be explained by several models (Section II). The Geneva Emotion Wheel \cite{88}, Differential Emotions Scale \cite{89}, and Appraisal Models are some of the models that use specific items to identify emotions depending on the events as they are framed in dimensional models (valence, arousal, and dominance). Analyzing emotional reactions utilizing multiple entities is crucial for grasping complicated emotions and capturing disparities in emotional responses. In light of this, it's crucial to examine how VR is able to evoke different types of emotions. Consequently, this section examines the potential use of various VR media formats in the induction of various behaviors.

\section{Datasets}
Our purpose in this section is to describe the datasets we use to detect emotions in text, audio, and visual images. There are three types of affective computing databases: textual, speech, audio, and visual. Models for emotional computing and network architectures are greatly influenced by these databases' characteristics.

\subsection{Text Based Dataset}
The TSA database contains text data at various granularities, like words, sentences, and documents. The Multi-domain Sentiment (MDS) \cite{90} database contains more than 100,000 phrases derived from Amazon.com reviews. These sentences are divided into two emotion types (positive and negative) and five sentiment categories. IMDB \cite{91} is another significant resource that is frequently utilized for categorizing binary sentiment. There are 25,000 reviews of extremely divisive movies available for testing and 25,000 for training. The Stanford University-annotated semantic lexical database is called Stanford Sentiment Treebank (SST) \cite{92}. It has 215,154 phrases with fine-grained emotional descriptors.

\subsection{Audio based dataset}
There are two types of speech databases: spontaneous and non-spontaneous (simulated and induced). The first non-spontaneous voice collection was created primarily from professional actors' performances. Due to their ability to professionally mimic well-known emotional traits, these performance-based datasets are considered trustworthy. There are over 500 words in the Berlin Database of Emotional Speech (Emo-DB) \cite{93} spoken by 10 actors (five men and five women) in various emotional states, including rage, anxiety, fear, boredom, and contempt. Artificial emotions, however, are more likely to be exaggerated than actual feelings. In an effort to close this gap, databases have been created for spontaneous speech.

\subsection{Visual base dataset}
To create the initial FER datasets, lab volunteers voluntarily expressed their emotions (in-the-lab). A 1998 publication entitled JAFFE \cite{70} contains 213 photographs of 10 Japanese female models expressing 7 different facial expressions. In order to develop the Cohn-Kanade+ (CK) \cite{94}, subjects were instructed to make 7 different facial expressions. In order to offer methods and benchmark findings for tracking features of face, action units (AUs), and identifying emotions, photos of face expressions were captured and evaluated. MMI \cite{96}, in contrast to CK, contains onset apex offset orders. Multiple views and multiple poses are supported by many 3D or 4D datasets made for FER. For the Binghamton University 3D Facial Expression (BU-3DFE) \cite{97} database, 606 facial expression segments were taken from 100 people using six different facial expressions.

A huge, unrestricted collection called FER2013 [98] contains 35,887 grey images with a resolution of 48 48 pixels that were autonomously gathered using the image search API 950,000 of the one million photos in EmotioNet \cite{99} were generated automatically. 91,793 facial photos from Emotion in the Wild (ExpW) \cite{100} have each been individually labeled using one of the seven fundamental facial expressions. The annotating technique eliminated non-face photos. Over one million facial photos are included in AffectNet \cite{101}, of which 4 lac 50k images have been manually classified as one of eight distinct expressions (neutral, disdain, the six fundamental emotions, and more). AffectNet also includes information on the dimensional intensity of valence and arousal. The Real-world Affective Face Database (RAF-DB)\cite{102} is a collection of 29,672 extremely different facial photos annotated by the public and acquired from the Internet (seven basic and eleven compound emotion labels). Over 16,000 video snippets that were cut from hundreds of films with different topics make up Dynamic-Facial-Expressions in the Wild (DFEW) \cite{103}.

\begin{figure*}[t]
    \centering
    \includegraphics[width=0.7\textwidth]{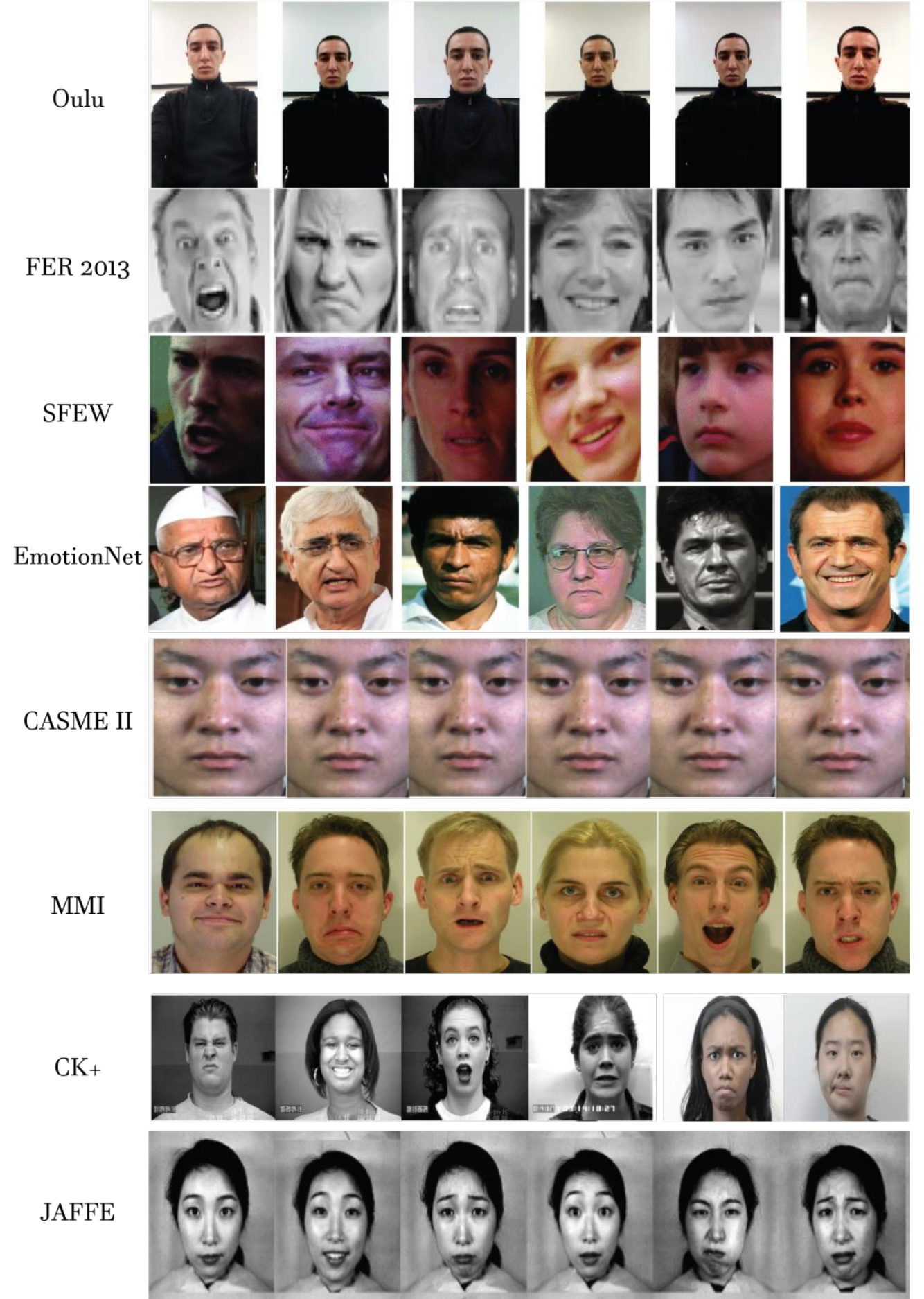}
    \caption{Sample images of openly available databases.}
    \label{fig4}
\end{figure*}

\begin{table}[h]
    \caption{Available databases for research purposes}

\centering
    \begin{tabular}{p{60pt} p{60pt} p{80pt} }
        \hline
        \hline
        \textbf{Database}&\textbf{Number of samples}&\textbf{Emotions} \\
        \hline
         Oulu \cite{104}& 80&SBE\\ 
        \hline
        
        FER2013 \cite{98}& 32&SBE\\ 
        \hline
        
        SFEW \cite{105}& 32&SBE\\ 
        \hline
        
        EmotionNEt \cite{99}& 32&Compound\\ 
        \hline
        Oulu \cite{104}& 32&SBE\\ 
        \hline
        
        ExpW \cite{100}& 32&SBE+\\ 
        \hline
        
       AffectNEt \cite{101}& 32 &SBE+\\ 
        \hline
        
        CASME II \cite{106}& 35s&SBE +\\ 
        \hline
        
        SMIC \cite{107}&16&3, positive, negatice, surprice\\ 
        \hline
        
        4DFAB \cite{108}& 180&SBE\\ 
        \hline
        
        3DFE \cite{97}& 100&SBE\\ 
        \hline
        
        MMI \cite{96}& 25&SBE\\ 
        \hline
        
        CK+ \cite{94}& 123&SBE+\\ 
        \hline
        
        CK \cite{95}&210&SBE+\\ 
        \hline
        
        JAFFE \cite{109}& 10&SBE\\ 
        \hline
        \hline
        \end{tabular}
    \label{tab5}
\end{table}

\section{Challenges}
ML-based techniques have been used in affective computing’s early literature \cite{17}. The ML pipeline \cite{110}entails the pre-processing of unprocessed signals, the creation of expertly constructed feature extractors (including selecting features when practical), and masterfully crafted classifiers. ML-based approaches for affective analysis are difficult to reuse across similar issues because to their task-specific and domain-specific feature descriptors, despite the fact that many types of hand-crafted features have been generated for diverse modality. The most popular ML-based classifier are the SVM classifier\cite{111}, GMM classifier \cite{63}, RF classifier \cite{112}, KNN classifier \cite{113}, and ANN classifier \cite{114}, with the SVM classifier being employed in the majority of ML-based affective computing tasks.

Due to their superior feature representation learning capabilities, DL-based models have recently gained popularity and outperformed ML-based models in the majority of affective computing applications \cite{115}. CNNs and their derivatives are made to extract significant and discrete characteristics from static information (such as face and spectral analysis images) \cite{116}. RNNs and its variants are made for capturing temporal dynamics for sequence information (such as physiological signals and movies) \cite{117}. The deep spatial-temporal feature extraction is a task that can be handled by CNN-LSTM models. By enhancing data and using cross-domain learning, adversarial learning is frequently employed to increase the robustness of models \cite{118}. Additionally, to enhance overall performance, various attention methods and auto encoders are used with DL-based techniques. The most discriminative traits appear to be automatically learned by DL-based algorithms, which appears to be an advantage. When compared to ML-based models, DL-based techniques have not yet had a significant influence on physiological emotion recognition.

In the field of emotion research, VR is being used more and more, but there are some limitations. Therefore, standards must be developed and implemented. Research can be conducted more effectively and restrictions can be reduced by applying these principles. It is important for researchers to assess the likelihood of different kinetic settings on the typical population in order to decrease motion sickness. By filtering participants at an early stage, it becomes easier to collect data. A questionnaire could be used as the primary method of contacting the individual.

A few of the questionnaires that can be found in the literature are the Motion Sickness Susceptibility Questionnaire (MSSQ) and the Virtual Reality Sickness Questionnaire (VRSQ) \cite{72}. It is possible for the reader to consult. Participants must have prior exposure to immersive VR environments, as with the first. Some research findings may be influenced by the novelty bias caused by using VR for the first time. Further explanations can be found in the differences between predicted and actual feelings.

People with previous VR experience or training sessions with selected participants before data collection can lessen the effects of the first-time VR experience. As a result of understanding the perspective of a VR world and the operation of hardware, especially when using VR controllers, participants receive additional benefits. For credible research, it is necessary to confirm the applicability of using those hardware interfaces integrated with different facial structures.

\section{Open Research Problems}
The field of affective computing focuses on developing machines and systems that can recognize, interpret, and respond to human emotions.
\begin{enumerate}
    \item    How to recognize and interpret complex emotional states more accurately and reliably is one of the open research problems in affective computing.
    \item     The lack of large and diverse labeled datasets to train emotion recognition models is one of the open research problems in affective computing. For ML algorithms to recognize and interpret emotions accurately, large amounts of labeled data are required. Existing emotion recognition datasets, however, are often limited in size, scope, and diversity, which can limit their performance and generalizability.
    \item     There is also a lack of consistency in the labeling of emotions. A standard labeling scheme for emotions does not exist, since emotions are interpreted differently by different cultures and individuals. It may be difficult to compare results across studies and develop emotion recognition models that are effective across diverse populations because of this.
    \item Another open research problem is how to develop more robust and interpretable ML models for affective computing. ML models used in affective computing, such as deep neural networks, are known as "black box" models, which make understanding how they work difficult. As a result, emotion recognition models are difficult to trust and fine-tune for specific applications due to their lack of interpretability.
    \item     A more personalized and adaptive emotion recognition model is needed that can take into account individual differences and changes in emotional states over time. Developing models that adapt to individual users' emotional patterns requires learning from their emotional expressions.
    \item     The development of ML approaches for affective computing and the development of more accurate and effective emotion recognition systems will be made possible by addressing these research problems.
\end{enumerate}

\section{Conclusion}
This survey examined research works based on current studies on affective computing and an illustrative taxonomy of affective computing. Various corresponding measures were used to assess the recognition outcomes obtained by classification or regression. Emotional computing develops computational models for DL-based or ML-based affective understanding, as well as benchmark databases for training.
A survey of the commonly used baseline databases for affective computing was also presented in this article, including textual, audio, and visual databases. Most affective analysis techniques can be applied to these publicly available databases. ML-based methods, DL-based approaches, and mixed reality-based approaches are some of the recent developments in affective computing. A survey of text sentiment classification, voice emotion recognition, and visual emotion recognition (FER and EBGR) is presented.
Even though affective computing systems that use either unimodal or multimodal data have made substantial progress, there are only a few reliable and potent algorithms to predict emotion and distinguish feelings. Consequently, affective computing's important future research directions are summarized as follows.
New and expanded baseline databases, particularly multimodal effect databases (textual, audio, visual, physiological), would be crucial.
Affective analysis problems such as FER under occlusions and fake emotion expression need to be solved.
Fusion methodologies, particularly those based on regulation or statistics, need to be improved.
In the presence of constrained or biased databases, zero/few-shot learning and unsupervised learning techniques, such as self-supervised learning, can improve the stability and robustness of affective analysis.
The use of emotive analysis in robotics is well known. As discussed in this paper, emotionally intelligent robots are capable of accurately mimicking and reacting to emotions.

\bibliographystyle{IEEEtran}
\bibliography{ref.bib}

\end{document}